# Automated Tracking and Estimation for Control of Non-rigid Cloth

Author: Marc Killpack
Advisors: Dr. Wayne Book and Dr. Frank Dellaert

# Table of Contents

# Preface (November 2012)

This report is a summary of research conducted on cloth tracking for automated textile manufacturing during a two semester long research course at Georgia Tech. This work was completed in 2009. Advances in current sensing technology such as the Microsoft Kinect would now allow me to relax certain assumptions and generally improve the tracking performance. This is because a major part of my approach described in this paper was to track features in a 2D image and use these to estimate the cloth deformation. Innovations such as the Kinect would improve estimation due to the automatic depth information obtained when tracking 2D pixel locations. Additionally, higher resolution camera images would probably give better quality feature tracking. However, although I would use different technology now to implement this tracker, the algorithm described and implemented in this paper is still a viable approach which is why I am publishing this as a tech report for reference. In addition, although the related work is a bit exhaustive, it will be useful to a reader who is new to methods for tracking and estimation as well as modeling of cloth.

# Introduction

Recent work on the modeling and tracking of cloth is focused primarily either on realistic recreation of cloth geometry and behavior or being able to simulate cloth in real-time. Research in the area of the fashion or movie industries has focused on the accuracy of cloth data capture in being able to accurately represent clothing and even replace it in post-production of a film [8][26][28]. Often this kind of work requires extensive setup, perhaps numerous cameras and does not run in near real-time. On the other hand, the emphasis in the video game industry is to model or simulate cloth in real-time. However most of these implementations or applications are focused on being aesthetically pleasing rather than physically accurate [30]. This means that although some aspects of cloth parameter estimation and tracking have become highly developed, the ability to control a piece of cloth using these models in real time with current methods is infeasible. In this work I developed an Extended Kalman filter cloth tracker that can run in near real-time (4-20 Hz) and accurately represent the state of the cloth for translations and rotations involving minor deformations. In addition I implement an EKF with a mesh model for the cloth, borrowing from finite element methods and show some promising initial results for tracking of large scale bending or folding in the cloth.

# Background

In this paper I describe my implementation of a tracking algorithm in order to track and control non-rigid cloth on a system with steerable conveyors, (as can be seen in Figure 1).

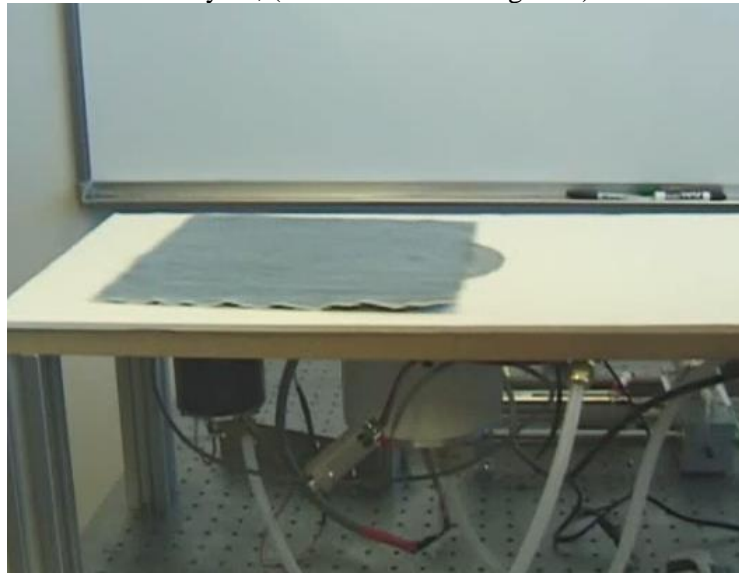

**Figure 1 Example of what a possible single steerable conveyor will look like.**

In our application we use two different models for the cloth. The first is the 2-dimensional x, y, and theta displacements and their derivatives of the center of mass of the cloth. The second model includes a 2-dimensional finite element mesh where the nodes represent the states of the cloth. The cloth to be used for this research is denim. Although for denim, the assumption of a rigid object for tracking will perhaps be sufficient, it is not general enough for other types of material or even extreme cases where the cloth moves rapidly or non-rigidly. For this reason a completely rigid model and a mesh model of the cloth will be considered.

Despite there being numerous methods for object tracking the technique of feature point tracking is most applicable for this project. This is because the cloth is non-rigid and the actuators in our system will require a model that can relate the states of the cloth to measureable feature points on the cloth.

We present the development of a cloth tracker using a mathematical cloth model, extracted 2D feature tracks from an image sequence and an Extended Kalman Filter formulation. This research extends current state of the art methods in non-rigid tracking and cloth modeling to a control application with the opportunity to help automate textile manufacturing in the future.

Future work should include completion of the automatic non-rigid cloth tracker by combining the more realistic cloth model and automatic 2D feature tracker in an Extended Kalman Filter. Implementation on the experimental setup for the steerable conveyors (which is an array of stepper motors, see Figure 1) will allow verification of accurate tracking and the development of a regulator scheme to keep the cloth flat. This work has provided the components for a viable non-rigid cloth tracking and control scheme. It has also extended current state of the art methods in non-rigid cloth tracking and modeling to a control application with the opportunity to automate textile manufacturing in the future.

## Overview of Cloth Tracking Algorithm

The tracking process involves four distinct events of 1) initialization, 2) state prediction, 3) measurement with data association and 4) state correction. The initialization stage concerns only the initial frames of the sequence. Background subtraction would be used to identify the cloth (foreground) from the background of the conveyor system except in the case where only the cloth is visible in the field of view of the camera. This work is done under the assumption of background subtraction for identifying the region of interest (ROI) and it is currently identified manually.

In order to effectively describe the algorithm three distinct spaces are defined. The first is the cloth space which is a two dimensional surface and is defined by variables s and t. The second space is the three dimensional real world space and is defined by Euclidean x, y and z coordinates. Finally, the third space is the image space and can be defined by u and v which describe the locations of pixels in a given frame during the tracking process. Space-mappings are defined such that we can move between the spaces.

Suppose that:

$$\boldsymbol{K} = \begin{bmatrix} f & 0 & 0 \\ 0 & f & 0 \\ 0 & 0 & 1 \end{bmatrix}$$

Where K is the ideal camera calibration matrix and f is the focal length. We can also say that:

$$\boldsymbol{M} = \begin{bmatrix} 1 & 0 & 0 & 0 \\ 0 & 1 & 0 & 0 \\ 0 & 0 & 1 & 0 \end{bmatrix} \times \begin{bmatrix} \boldsymbol{R} & \boldsymbol{t} \\ 0 & 1 \end{bmatrix} = \begin{bmatrix} 1 & 0 & 0 & 0 \\ 0 & 1 & 0 & 0 \\ 0 & 0 & 1 & t_z \end{bmatrix}$$

Where R is the camera rotation matrix with respect to the global coordinate system and t is the translation of the camera away from the global coordinate system. Let *u* and *v* represent the pixel coordinates in image space then the following is true:

$$\begin{bmatrix} u \\ v \\ 1 \end{bmatrix} \sim K \times M \times \begin{bmatrix} X \\ Y \\ Z \\ 1 \end{bmatrix}$$

This leads to:

$$\begin{bmatrix} X \\ Y \\ Z + t_z \end{bmatrix} \sim K^{-1} \begin{bmatrix} u \\ v \\ 1 \end{bmatrix} = \begin{bmatrix} x \\ y \\ z \end{bmatrix}$$

Since the cloth is assumed to be flat in the initial frame the scalar which satisfies the following can be approximated:

$$Z + t_z = \alpha z$$

$$\alpha = \frac{t_z}{z}$$

Then we make the approximation that:

$$\begin{bmatrix} X \\ Y \\ Z \end{bmatrix} = \begin{bmatrix} \alpha x \\ \alpha y \\ \alpha z - t_z \end{bmatrix}$$

All of this can be done to give the 3 space coordinates of the original feature points which we represent as follows:

$$(x, y, z) = w(s, t)$$

The mapping from the real world space to the camera space can be defined as:

$$(u, v) = \Pi(x, y, z) = \Pi(w(s, t))$$

This is essentially the inverse of the mapping from image space to world space and can be found as follows:

$$\Pi(x, y, z) = K \times M \times \begin{bmatrix} X \\ Y \\ Z \\ 1 \end{bmatrix}$$

Where K and M are the camera calibration matrices defined above.

The definition of these spaces is necessary because the measurement is done in the image space whereas the object exists in real space and the features exist in cloth space and the three must be related. State prediction and correction steps will take place in the real world space. For the rest of the algorithm these mappings remain defined as above except if an estimate of Z in world space were available in which case this would slightly modify the approximation of alpha in the development above.

During the initialization stage the cloth position is set in both cloth and real-world space coordinate systems.

In the first frame "features" within the perimeter of the cloth are identified in the image using feature extractors. Those points are projected into the cloth space. If **q** is a 2 by n matrix containing the pixel coordinates of the n features, then the cloth coordinates can be found as follows:

$$(s, t) = w(q_s - 320, 240 - q_t)$$

Where w is the mapping between image and real world space and the values of 320 and 240 are used to center the image coordinates at zero for a 640x480 pixel image.

These points are then projected onto the cloth space or coordinate system where essentially:

$$\begin{bmatrix} s \\ t \end{bmatrix} = \begin{bmatrix} X \\ Y \end{bmatrix}$$

This being completely valid only for the initial frame unless we assume that the cloth always behaves as a rigid plate. Once these points are identified and mapped to the cloth space, the initialization stage is completed.

The rest of the process is a repeating cycle which can be seen as follows:

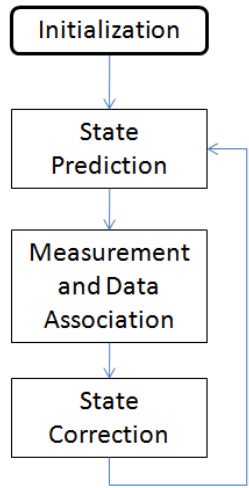

The cloth has been represented in one of two ways. In initial development the cloth was modeled as a rigid object for simple validation purposes and has since been replaced by the mass-spring model developed by Provot [50]. The cloth is assumed to be flat in the initial frames and the camera calibrated. For the mass-spring model a square mesh is mapped onto the cloth. The nodes of this mesh represent the state of the cloth.

The prediction of the states of the cloth at each time step is done using the Nvidia physics engine called PhysX. Once the initial mesh and cloth parameters are defined, a prediction can be made for a given time step using only the previous locations and an estimated applied force.

Measurements are taken from the current frame in order to re-identify the locations of the originally identified feature points whose coordinates were stored in **q**$_o$ in the initialization stage. We associate the newly detected feature points with the points that have been tracked up until **q**$_{k-1}$ at time k.

After having resolved the data correspondence problem in the current step, the original estimate of the states at the current time step can then be corrected.

For the formulation of the Extended Kalman Filter we follow the formulation by Welch and Bishop [71]. The Jacobian matrix J of the homogenous state matrix is calculated using complex step differentiation. This numerical Jacobian is a mxm matrix where m is the number of states.

We can then calculate our a priori estimate error covariance as:

$$P_k^- = JP_{k-1}J^T + Q$$

Q is defined as the covariance matrix for the process noise and the integer k represents the current time step.

For the two different cloth models (rigid and mesh model) there are two different measurement functions. The measurement function for the rigid model involves simply rotating the original set of feature points **q** in cloth space by the current estimate for theta and then translating them.

$$\boldsymbol{q}_{x,y,z} = \begin{bmatrix} \cos(\theta) & -\sin(\theta) & 0 \\ \sin(\theta) & \cos(\theta) & 0 \\ 0 & 0 & 1 \end{bmatrix} \begin{bmatrix} \boldsymbol{q}_{s,t} \\ 0 \end{bmatrix} + \begin{bmatrix} X \\ Y \\ 0 \end{bmatrix}$$

Where **X,Y**, and, θ are three of the estimated states of the cloth at each step.

Using the same K and M matrices that were defined earlier the following defines the measurement function given the current prediction of the states X,Y, and θ for each feature point *i*:

$$h_{step}(\boldsymbol{q}) = \boldsymbol{K} \times \boldsymbol{M} \times \begin{bmatrix} \boldsymbol{q}_{x,y,z}^i \\ 1 \end{bmatrix} = \begin{bmatrix} \alpha u_i \\ \alpha v_i \\ \alpha \end{bmatrix} \quad for\ i = 1,2,\dots n$$

$$h_i(X,Y,\theta,\boldsymbol{q}) = \begin{bmatrix} \dfrac{u_i}{\alpha} \\ \dfrac{-v_i}{\alpha} \end{bmatrix} + \begin{bmatrix} 320 \\ 240 \end{bmatrix}, \quad (where\ the\ image\ is\ 640x480)$$

The measurement function for the mesh formulation uses the concept of shape functions from finite element methods. The general idea is to weight the corners of a mesh square according to where the feature point is found within that square. Each square in the mesh is initially mapped to a square centered at zero and with a width and length of two. This can be seen in Figure 2.

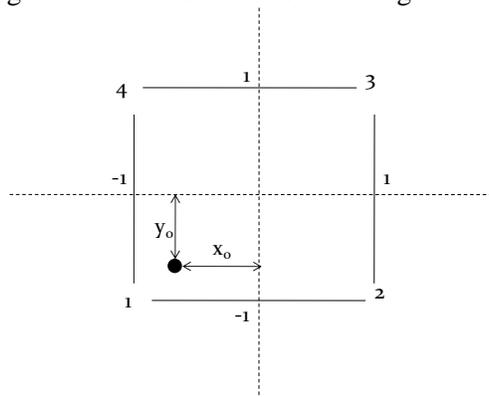

**Figure 2 Schematic of mapping from original mesh to square used for shape functions in FEM.**

The corners of the square are then weighted according to the location of the feature point within the square as follows:

$$N_1 = \frac{1}{4}(1-x_o)(1-y_o)$$

$$N_2 = \frac{1}{4}(1+x_o)(1-y_o)$$

$$N_3 = \frac{1}{4}(1+x_o)(1+y_o)$$

$$N_4 = \frac{1}{4}(1-x_o)(1+y_o)$$

This allows estimation of the location of the feature at any time with only the state estimates that are calculated anyway. A representation of this mapping can be seen below in Figure 3.

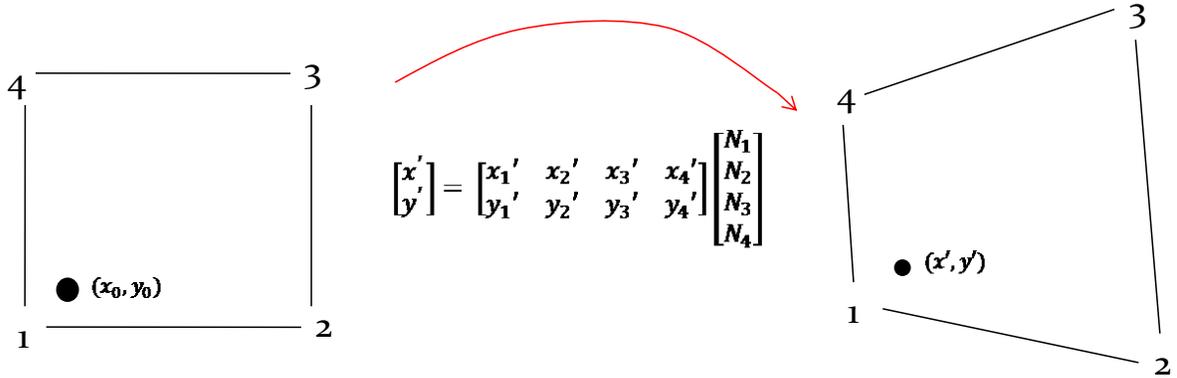

**Figure 3 Mapping from original mesh to location of new coordinate still relative to original nodes.**

Given one of the two described measurement methods, the Jacobian of the measurement function can now be calculated as the state variables are varied and call this matrix F. As each measurement is two dimensional (u and v coordinates) we stack the total measurement vector as a single column vector of all the u's and then the v's. This means that the measurement vector is 2*n where n is the number of measurements. This also means that F is 2*n by m where m is the number of states.

The gain matrix which weights the residual error between the measurement function and the actual measurements (W) is then found as follows:

$$K_k = P_k^- F^T (F P_k^- F^T + R)^{-1}$$

The current state estimate is then calculated by using the original prediction for the time step and adding the quantity of the gain matrix times the residual:

$$\widehat{X_k} = X_k^- + K_k(W - h(X, Y, \theta, \boldsymbol{q}))$$

Where W is the actual measurement performed by the 2D point tracking algorithm. Finally, the a posteriori estimate error covariance can be calculated:

$$P_k = (I - K_k F)P_k^-$$

At this point the next prediction is made and stages two through four of the algorithm are repeated in an effort to continuously track the non-rigid cloth. There are obvious limitations as the model and tracking will likely fail for extreme deformations or tangling of cloth, but the assumption of application is extremely important. As this tracking method is to be used in a control scheme, initial wrinkling or folding that is detected should be smoothed by the actuators.

# Development, Testing, and Results for Part 1

The work described in this section is the result of the first semester of a two semester research course.

## Experimental Setup

My cloth tracking algorithm was initially implemented in Matlab and three cloth movement tests were chosen. The first is the simple translation of the cloth in a single direction with one applied force in that direction. The second test is a rotation and translation of the cloth induced by a force with a moment arm. The third test is the compression and tension of the cloth from both sides causing a folding and unfolding in the middle. These tests permit us to test the limits of our current implementation and look for improvements.

In order to test the basic procedure as previously noted, the cloth in the image as well as twenty features were manually identified across a certain number of frames containing motion of the cloth. Since feature extraction is a fairly developed field, we focus instead on the cloth tracking problem. We use an idealized camera calibration matrix and set the camera directly above the cloth in order to make assumptions about R and t in the camera matrices as described above. The cloth was filmed using a 640x480 resolution camera at 30 fps. This data was then imported into Matlab. For the real application of control, this would obviously need to be done in real-time.

## Results and Discussion

For each of the three tests described above the same variations were used. Once the frames were read into Matlab, we ran the algorithm with:

-no assumed model or force
-only the assumed force
-an assumed force and the Extended Kalman Filter for the rigid model only
-no assumed force and the Extended Kalman Filter for the rigid model only
-an assumed force and the Extended Kalman Filter for the mesh model
-no assumed force and the Extended Kalman filter for the mesh model

The results for each part and each test are not included below. However, those thought to illustrate a point or give insight into the performance of our algorithm have been included. In the frames showing the algorithm's progression, the green x's are the predicted features, the red x's are the measured ones and the yellow line is the error between them. For the error graphs also shown below, we used the magnitude of the distance between the predicted feature location and its measured location to quantify error. Since the measured features were extracted manually, this seemed to be a logical baseline for error comparison.

# Y-direction applied force

The images in Figure 4 are for no assumed force or cloth model and show how the error progresses from the 1st, 20th and 30th frames.

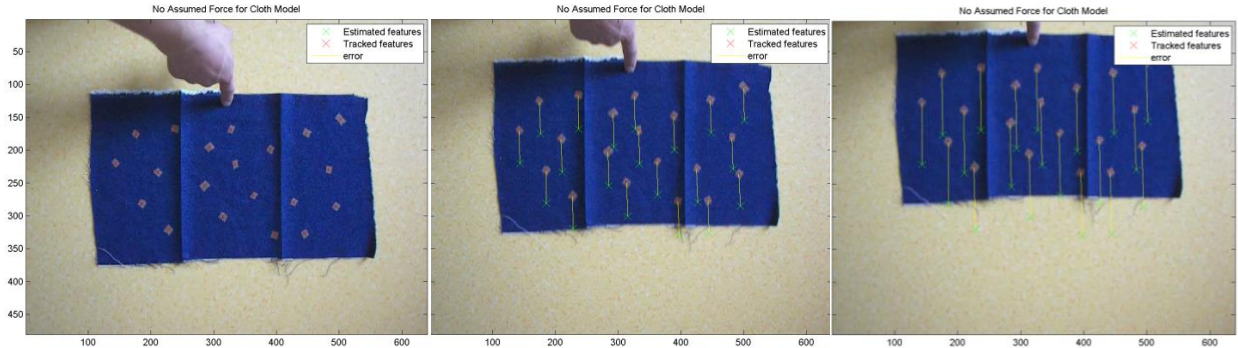

**Figure 4 Frames 1, 20 and 30 for sequence with no assumed cloth model or force.**

The images in Figure 5 shows the progression of the algorithm for the case where the Extended Kalman Filter is used with the rigid cloth model.

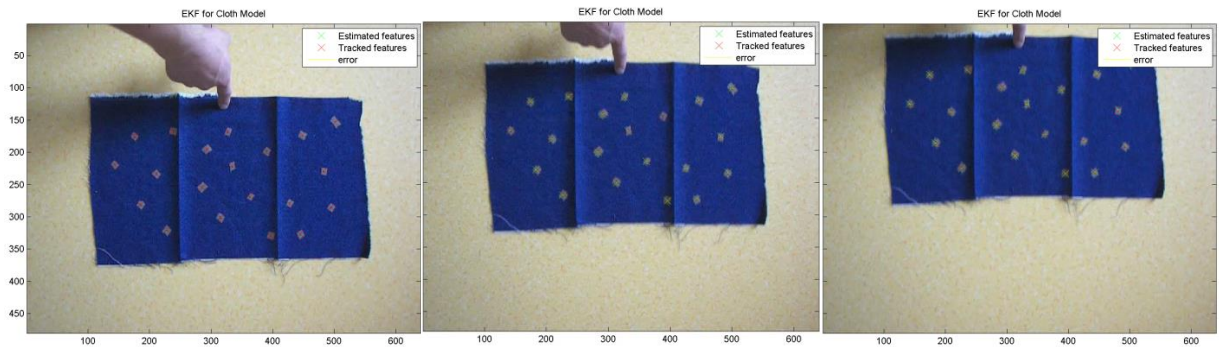

**Figure 5 Frames 1, 20 and 30 for the sequence using the EKF rigid cloth model implementation.**

The error graphs below show the comparison when only the force is assumed versus when we use the Extended Kalman Filter. This intial result is encouraging as we have only a single pixel average error for the twenty measurements over most of the frames in the sequence for the EFK implementation.

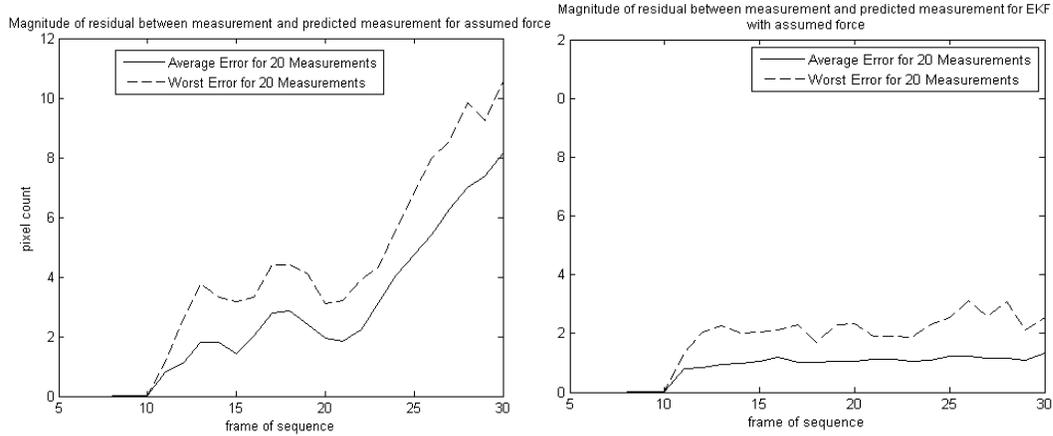

**Figure 6 Left: error for assume force only model. Right: Error for EKF rigid cloth model.**

## Applied Moment:

The images in Figure 7 again show the progression of the cloth and the error through the sequence of frames with no assumed force or model.

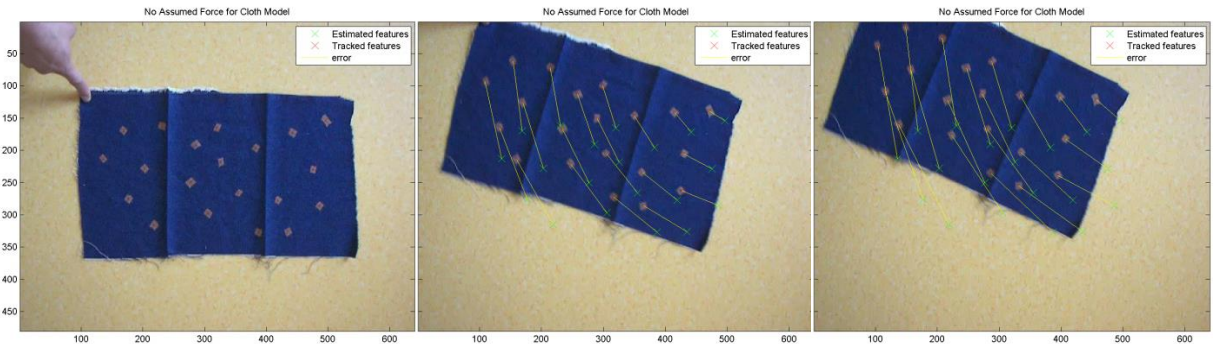

**Figure 7 Frames 1, 20 and 30 for sequence with no assumed cloth model or force.**

In Figure 8 top and bottom we can see the progression of the algorithm using the mesh model and the EKF. For the first figure we have assumed an applied force that is not included for the second. As the nodes of the mesh represent the states of the cloth, it is clear that the second figure, which assumes no applied force, is qualitatively less accurate.

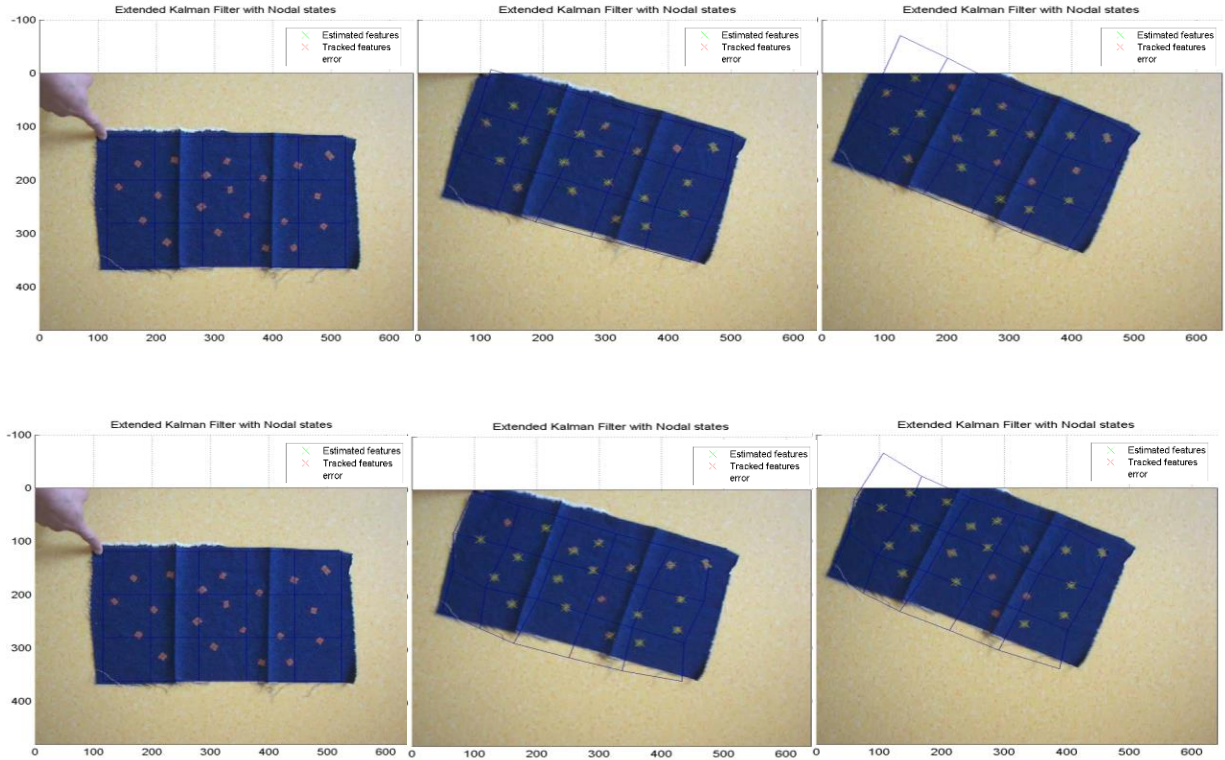

**Figure 8 Above: EKF mesh model with assumed force, Below: EKF mesh model with no assumed force for frames 1, 20 and 30.**

The error graphs in Figure 9 make it clear that for the rigid cloth model, the EKF implementation with an assumed force easily outperforms the one with no assumed force. However, although obvious by observation in the frames above, the error graphs make the EKF mesh model with and without force look comparable. The reason for this is because we are only comparing the error in the measurements and not in the actual state estimates. The main limiting factor in our algorithm is that although the shape functions for the mesh currently work to push the nodes or states into positions so that the measurements will match, there is no physical phenomenon keeping the nodes in likely relationships with each other. This is abundantly clear in the next test.

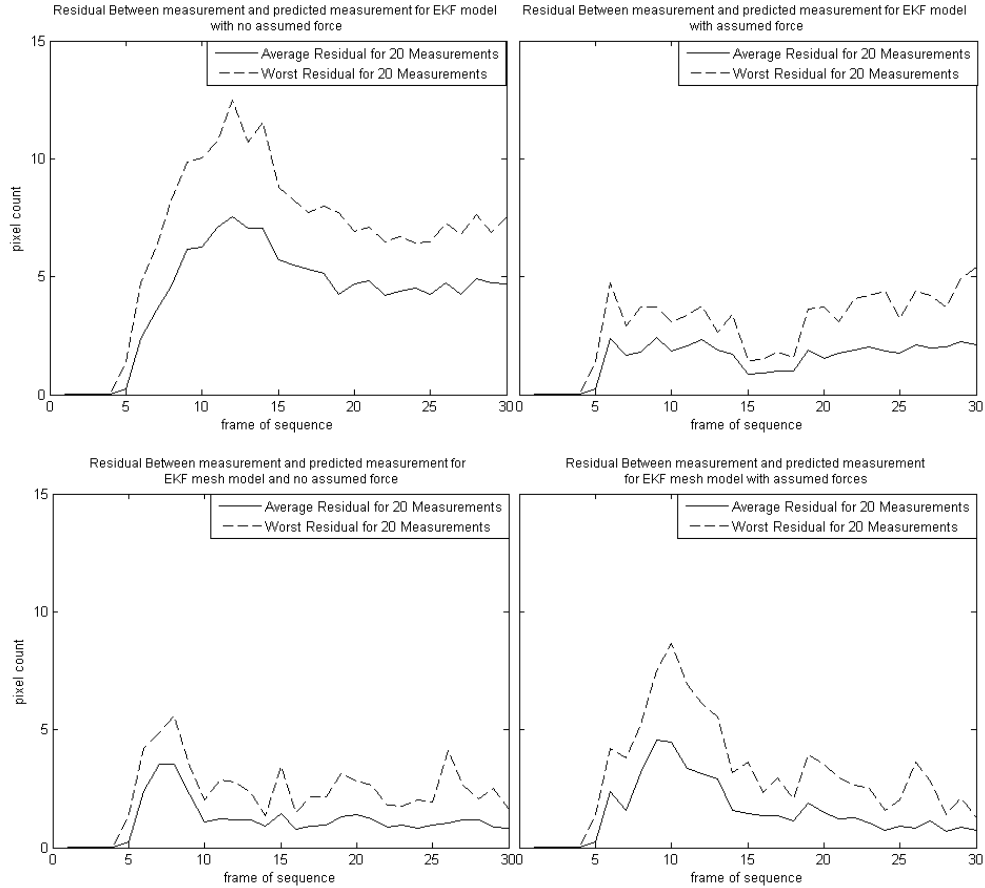

**Figure 9 Top left: Error for the EKF model with no assumed force. Top Right: Error for EKF model with assumed force. Bottom left: Error for EKF Mesh model with no assumed force. Bottom right: Error for EKF mesh model with assumed force.**

## Applied Compression and then Tension :

In this test we applied an approximately equal force to each side of the cloth first pushing and then pulling it back into place. The frames shown below were frames 1, 30 and 45. This means that the entire sequence occurs within a second and a half since the camera is filming at 30 frames per second. Figure 10 again simply shows the progression of the cloth and the error without any model or filter for the given sequence.

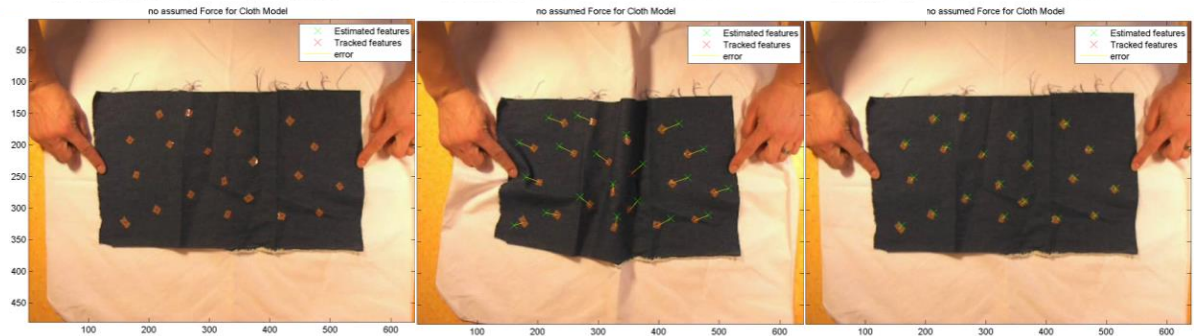

**Figure 10 Frames 1, 30 and 45 for sequence with no assumed cloth model or force.**

Figure 11 is very telling as to the limitation of our current algorithm but also shows its potential. Initially we can see that the mesh deforms rather absurdly and does not even return to its original shape as the cloth is brought back to its initial position. However upon further examination, one can see that the middle square and some squares around it deform quite accurately with the bending of the cloth.

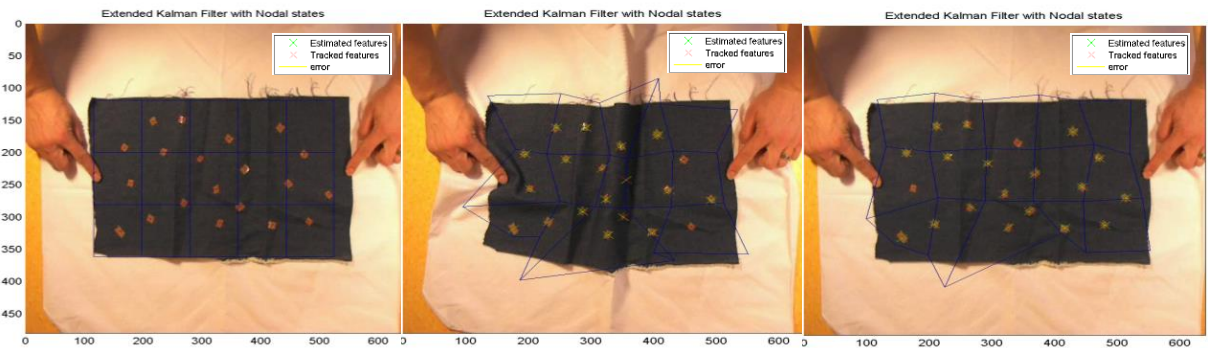

**Figure 11 Frames 1, 30 and 45 for sequence with EKF mesh model**

Furthermore looking at the graph of error shows that although the nodes are completely wrong in estimating many of the states, the measurement error is only large when the cloth initially changes direction as the force is applied. The reason for this is that the shape functions and measurement weighting in the extended Kalman filter are working correctly. They are pulling the states into positions which minimize the error in the measurement. In the case of the middle square there are enough measurements that it is actually constrained to move in approximately the right way. However for all the other nodes they are lacking a physically realistic weighting to keep them close to the other nodes or at least moving together. The other two error graphs show that even in the case of bending, the rigid model more accurately models the bending or overall motion than no model  However,  this is not good enough for control of the cloth in cases where folding and bending are likely to happen for larger pieces of cloth or less stiff pieces.

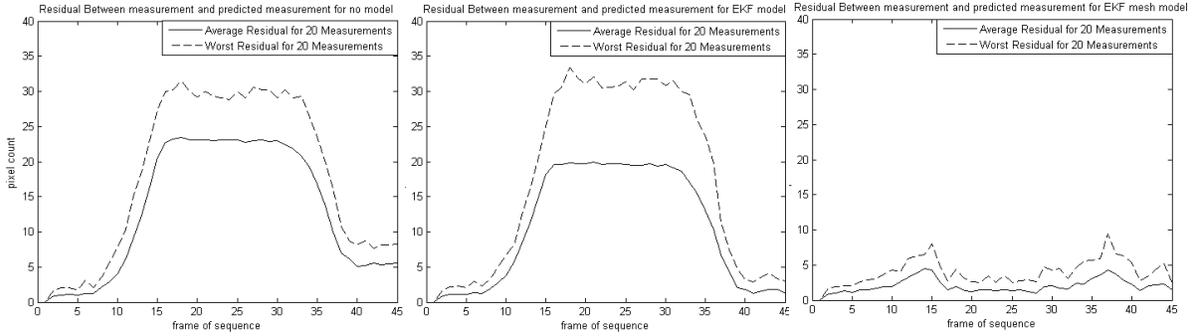

**Figure 12 Error for no model, EKF rigid model and EKF mesh model.**

What is required in order to improve the algorithm is a simple physically realistic weighting for the nodes of the cloth. This could be accomplished by geometrically limiting the nodes to not move farther apart then their initial relaxed state. However a mass-spring model which is currently being implemented is more likely to give accurate predictions for the cases of compression and folding.

Table 1 below shows the run-time per frame that was required for recent work in this area by other authors. Although our algorithm is not currently running in real-time, it is a vast improvement over past work. In all fairness, their objectives were not real-time or control, however using the same basic principles of tracking features and using cloth models we were able to implement a much quicker algorithm.

**Table 1 Comparison of run-times for different algorithms.**

| Authors | Rate of analyzing frames |
|---|---|
| Pritchard et al. (2003) | 0.0028 fps |
| Hasler et al. (2006) | 2.7E-5 fps |
| Hernandez et al. (2007) | 0.025 fps |
| Bradley et al. (2008) | 2.78E-4 fps |
| Killpack et al. (2008) | 4 fps |

# Conclusion for Part 1

I have presented a method for modeling and tracking the non-rigid motion of cloth and made some important steps towards a viable solution for cloth control. Particularly important is the recognition that using shape functions from finite element theory has allowed us to effectively weight our measurement estimates and therefore find the minimum error in measurement. What is obviously necessary to complete the work and make it successful, (more than just a rigid cloth tracker), is a more robust and realistic physical model of the cloth and the nodes representing the cloth.

Future work will include an immediate implementation of the mass-spring cloth model developed by Provot [50]. In addition, using faster 2D point trackers (see [61]) will allow us to extract and track around a thousand features at 30 fps. This will also improve the robustness of our algorithm. Another aspect that will improve the work would be to run a simple optimization scheme (something like a genetic algorithm or perhaps direct search) to find the parameters for spring constants for the mass-spring model that will give the least amount of error for a baseline sequence of the cloth moving. Finally the code can be optimized and embedded in order to attain real-time cloth tracking and the control law to control the cloth on our given conveyor system can be developed.

# Development, Testing, and Results for Part 2

The work described in this section is the result of the second semester of a two semester research course. Proof of concept had been shown from results in Part 1. This section outlines results relating to improvements in the cloth model and implementation for tracking using the LabVIEW software and camera.

## Experimental Design and Simulation

*Cloth Model*

Originally a two degree of freedom mass-spring model was formulated in Matlab according to the model in [50]. This was done as a replacement for a rigid plate cloth model. The choice of two degrees of freedom was an attempt to reduce the computation time and complexity of the algorithm. By assuming the cloth deformed in the plane, the z-component could be ignored and collision detection with the ground (which can be computationally costly) could be avoided. However upon initial implementation and experimentation it was seen that a non-linear spring model would be required in order to accurately predict the motion of the cloth. Figure 13 shows a result that was not unexpected. Due to the nature of numerical integration and since the cloth is constrained to the plane, the resulting movement of the cloth was a total acceleration either downwards or upwards when put in compression in the x-direction.

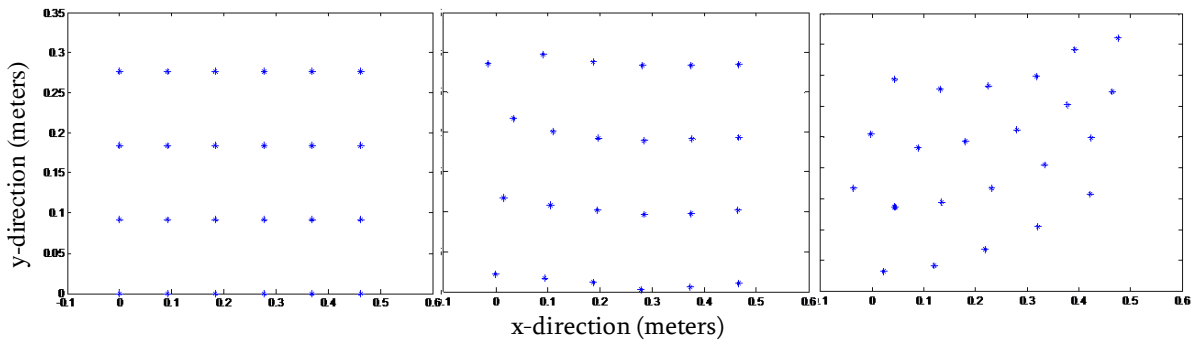

**Figure 13 Deformation and translation of representative mass nodes of a cloth when in compression.**

In addition to being inaccurate, this initial implementation was also computationally slow. Since the cloth and the differential equations governing the cloth are both stiff, the ODE solver in Matlab was extremely slow. Such that, even if a non-linear spring model was developed, the overall mass-spring model would have required implementation in a different programming language in order to run at acceptable frame rates.

These initial results led to an implementation of Nvidia's physics engine as a viable alternative. This is preferable for a number of reasons. The first is that the code for PhysX (the commercially developed physics engine for Nvidia) has already been highly optimized to run in real-time and includes collision detection and three degrees of freedom in the model. Secondly, it permits a user-defined mesh size and simulation time step as well as the ability to change numerous cloth parameters such as damping and stiffness in order to match physical cloth behavior. In addition to position variables being returned for the mesh nodes, the program can also estimate velocities which could be useful for control. Finally, it allows the representation of the conveyors which will control the cloth as spherical force fields rather than forces acting at specific nodes of the cloth mesh.

In order to evaluate the effectiveness of the tracker a brief discussion of how error will be measured is necessary. Since the nodes of the cloth are imposed and not true physical locations on the cloth, it is difficult to measure the actual error between the states and their estimates. An alternative is instead to measure the error between measured features on the cloth and their estimated locations as given by the estimated states. In order to assign an actual value with the error of the tracking method, the absolute distance between the measurement's pixel values and its estimate are calculated. In initial tuning, features can be manually identified and this measure can then be accurately termed error. However, in subsequent testing with the automatic 2D point tracker, this measure is referred to as pixel residual since it is the difference between the measured and predicted features. The average is recorded as well as the worst pixel error for each frame. As the distance error is already an absolute measure, there is no need for an RMS sort of calculation. One final note is that although the error between manually extracted feature locations and their estimates in the image is a generally effective measure, it is not definitive. Early tests have already shown that an observer can see estimates of states that give good feature estimates but very poor error qualitatively (see Figure 14 below). This means that at least in initial testing, in addition to the average pixel error and the worst pixel error there will need to be a physical observation of the estimates of the state to verify accuracy. Further work could involve using another calibrated camera as a static observer from a different angle to verify the estimated states of the cloth.

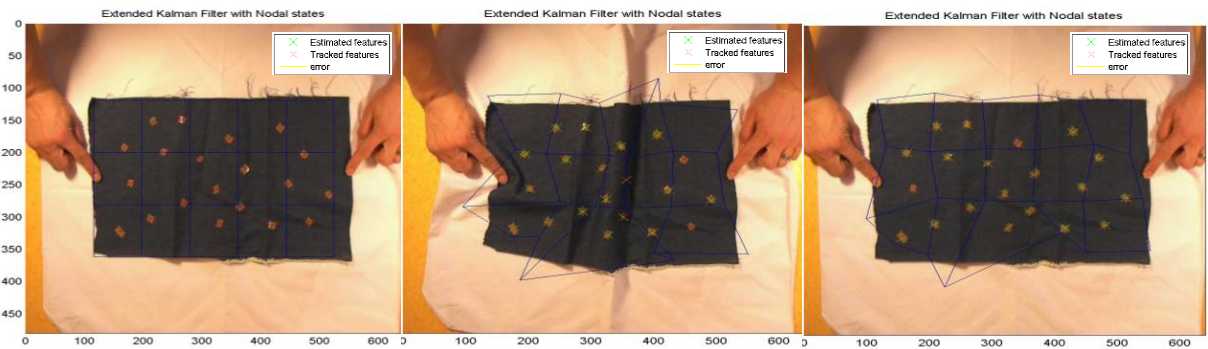

**Figure 14 Example of state estimates that give good measurement estimate but are qualitatively poor.**

The first step in developing the new cloth model was to tune the parameters of the physics engine cloth model in order to match as much as possible the behavior of the denim cloth to be tracked and controlled. Parameters of dynamic and static friction coefficients, stiffness, damping and bending are all tunable in the PhysX SDK. Using a set of video sequences typical of cloth motion (translation, rotation, bending and folding), a set of manually extracted features for each frame was used to minimize the measurement estimate error. The average pixel error for a sequence of frames was weighted with the worst pixel error and combined as a cost function. This was done, because although we want to reduce the worst error over all of the frames, it is also important to have a generally low overall average error.

This cost function was minimized using a genetic algorithm [29] that varied the parameters of the simulation to change the value of the cost function as follows:

$$max \ f(x) = -[(average\ pixel\ error) + 2 \times (worst\ pixel\ error)]$$
$$s.t. \ a_i < x_i < b_i \ , \quad \forall \ i = 0 \ldots n, \quad for\ n\ variables$$

Where f(x) is the cost function, $x_i$'s are the cloth parameters and $a_i$ and $b_i$ define reasonable constraints on the parameters. Although genetic algorithms do not necessarily find a global optimum, they have been shown to effectively search large and complex solution spaces to find acceptable solutions. An acceptable solution in this case is one that reduces the simulation error in pixel measurement estimate to values that correspond to noise levels in the measurement technique (on the order of 1-2 pixels). As can be seen in Figure 15, error values near one pixel for a sequence of cloth translation were easily obtained.

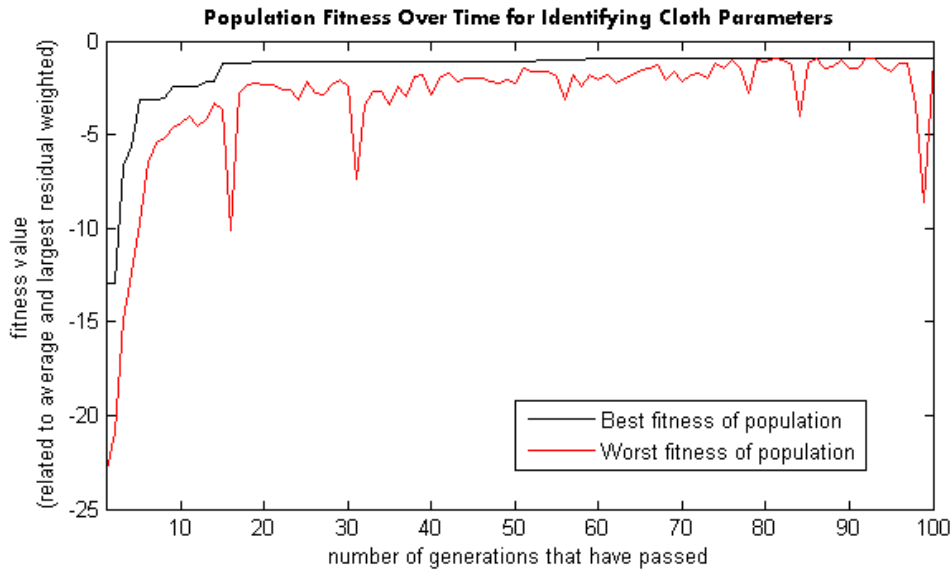

**Figure 15 Example of parameter identification using a genetic algorithm on a cloth translation video sequence.**

In addition to the simplicity of implementing genetic algorithms, another benefit of using an evolutionary type algorithm is that the result is a population of solutions. This allows the user or engineer to make practical and physically meaningful decisions based on a population of best designs. One example of this principle can be seen for the cloth model in Table 2 below. The values are for the best design in the given generation. Nvidia's PhysX engine only gives some parameters as a value from 0 to 1 which is why gravity was added as a variable to give more flexibility in the model.

**Table 2 Parameter and fitness values for two different generations.**

|  | Generation 87 | Generation 89 | Lower Limit | Upper Limit |
|---|---|---|---|---|
| Bending stiffness | 0.94 | 0.26 | 0 | 1 |
| Stretching stiffness | 0.74 | 0.74 | 0 | 1 |
| Density of Cloth | 0.036 | 0.036 | 0.02 | 2 |
| Thickness of Cloth | 0.0032 | 0.0032 | 0.0005 | 0.01 |
| Damping of Particles | 0.49 | 0.49 | 0 | 1 |
| Iterations of Solver | 4 | 4 | 1 | 5 |
| Friction Coefficients | 0.079 | 0.079 | 0 | 1 |
| Gravity Value (to induce stiffer cloth) | 12 | 12 | 5 | 15 |
| Fitness Value | -0.89 | -0.89 | | |

The two designs from different generations given in Table 2are almost identical except for the Bending Stiffness value. Their fitness values differ only in the fifth significant digit. However, generation 89's value for bending stiffness being so low would be physically inaccurate for denim when the cloth is put in tension. This "family" of solutions allows the use of good physical judgment in addition to the optimization routine.

*Cloth Results*

Using the genetic algorithm and manually identified features, the cloth model was tuned to give acceptable performance. Of particular importance was the ability of the cloth model to predict bending and folding of the cloth. In Figure 16, on the left is the image of the real piece of denim in compression. On the right is the simulated piece of cloth that was solved for using Nvidia's physics engine and displayed using OpenGL.

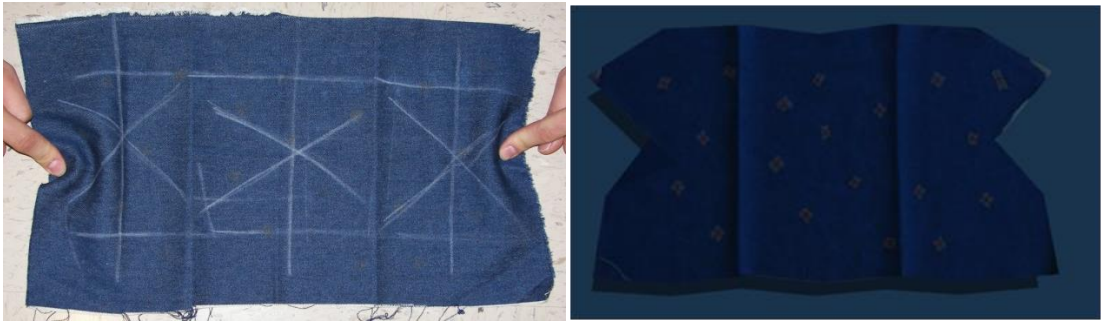

**Figure 16 Bending deformation of cloth in compression, on the left is real image; on the right is simulated image using OpenGL and Nvidia physics engine.**

It is important that for different types of cloth or thickness of cloth, the model must still be accurate. For this reason, another video was taken using a thicker piece of cloth placed under the first. In Figure 17, it can be seen that it was also possible to simulate the real physical behavior.

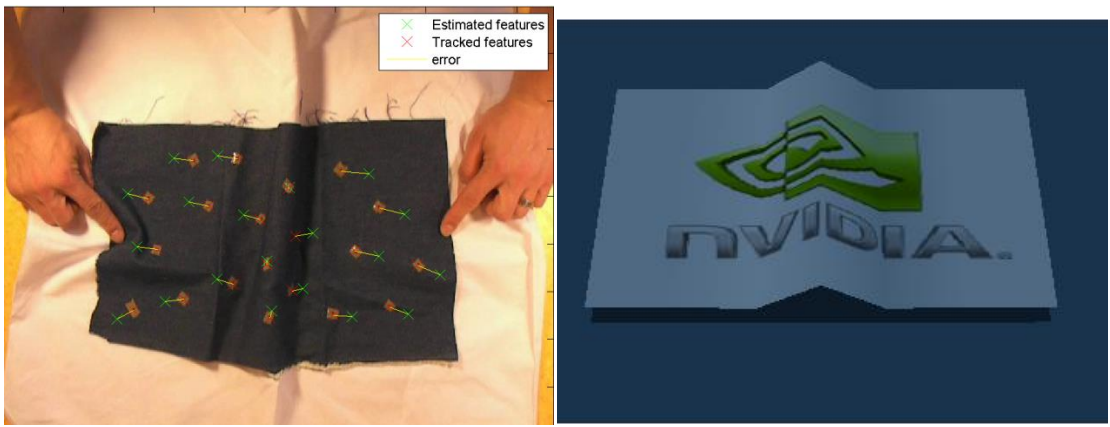

**Figure 17 Real and simulated bending deformation of a thicker piece of cloth.**

These two figures, although not conclusively, show the specific ability of the cloth model to accurately predict certain motions with given force inputs. The image on the left in Figure 17 shows the error that was associated when tracking with the previous rigid plate model for the cloth. This new cloth model should address this issue sufficiently well.

As speed was an issue for the initial Matlab implementation of the cloth model, this Nvidia cloth model was tested for speed. On an Intel Quad-core 2.33 GHz desktop computer, the model was able to simulate one time step of $1/30^{th}$ of a second at or under 15 ms. This is an acceptable speed for the cloth model.

*2D feature tracking*

The next stage was to implement a 2D point tracker. Although the GPU KLT tracker [61] purportedly tracks near 1000 features at around 30 fps there were two major reasons to not use their method in this application. The first was that it required a specialized and expensive Nvidia or ATI GPU. In addition, although it claims to track 1000 features, most of their demo work showed only an average of around 300 features being tracked. One important note however, is that using the GPU unloads the CPU and would allow a speedup for the total cloth tracker system. For this reason, using the GPU KLT tracker will be considered in future work if this initial implementation of the cloth tracker is considered successful enough.

Instead, the OpenCV implementation of Bouguet's work [6][9] was used to do pyramidal Lucas-Kanade tracking. Following the general OpenCV implementation, Shi and Tomasi's work [60] was used to find the initial features that were then tracked from frame to frame.

This algorithm, as well as the cloth model simulation was compiled into dynamically linked libraries in order to communicate with a National Instruments Smart Camera. The architecture for the experimental setup is shown below in Figure 18.

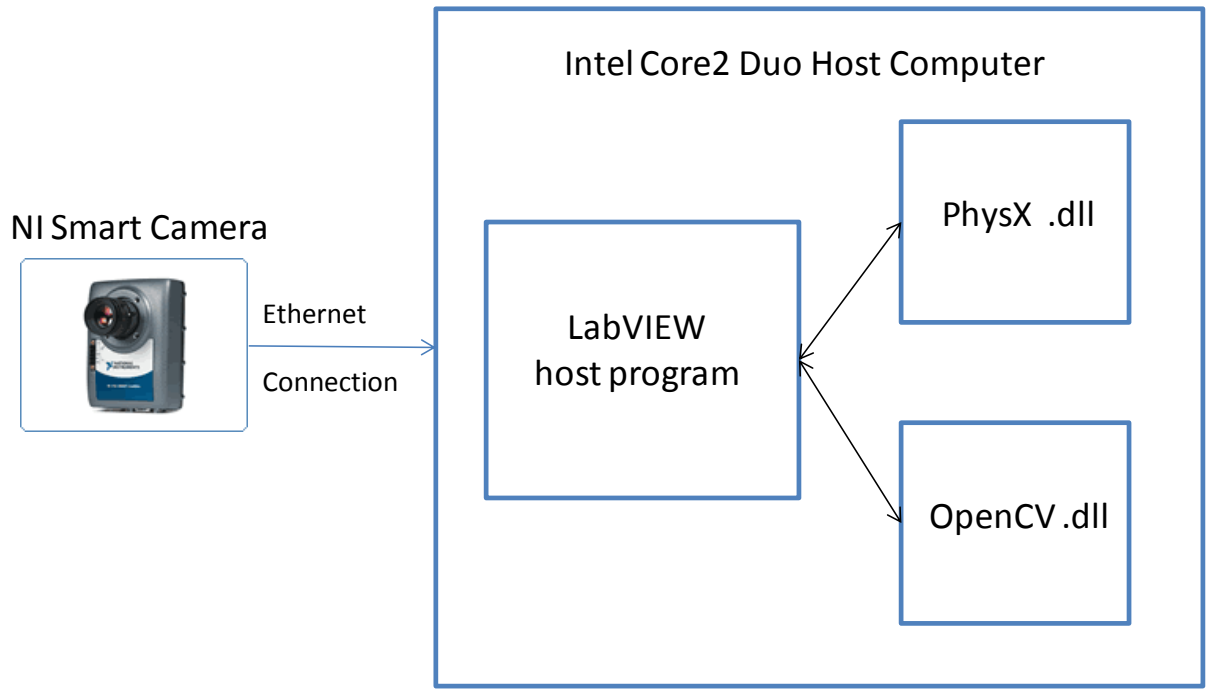

**Figure 18 Architecture of experimental setup for cloth tracking tests.**

The Smart Camera from LabVIEW can take 640x480 resolution images at between 30 and 60 fps. Each frame can then be passed back to a host computer where a LabVIEW host application is running and handles communication with the camera. The current and previous frames as well as a 2D array of features from the previous frame are then sent to the OpenCV .dll. The OpenCV .dll does pyramidal Lucas-Kanade tracking and returns the tracked feature locations to the host program. The PhysX .dll meanwhile, takes the previous state estimates and any inputs from the stepper motors and passes back a prediction of the state estimates at the next time step. The rest of the cloth tracking algorithm proceeds as outlined in the Appendix.

*Initial Results from Smart Camera*

In order to test the 2D point tracker implementation, translation and then a shearing motion were used to check the accuracy of the tracker. Initial results were encouraging. Figure 19 shows the feature tracking working in real-time with the NI smart camera between 10-15 fps when a shearing force is applied.

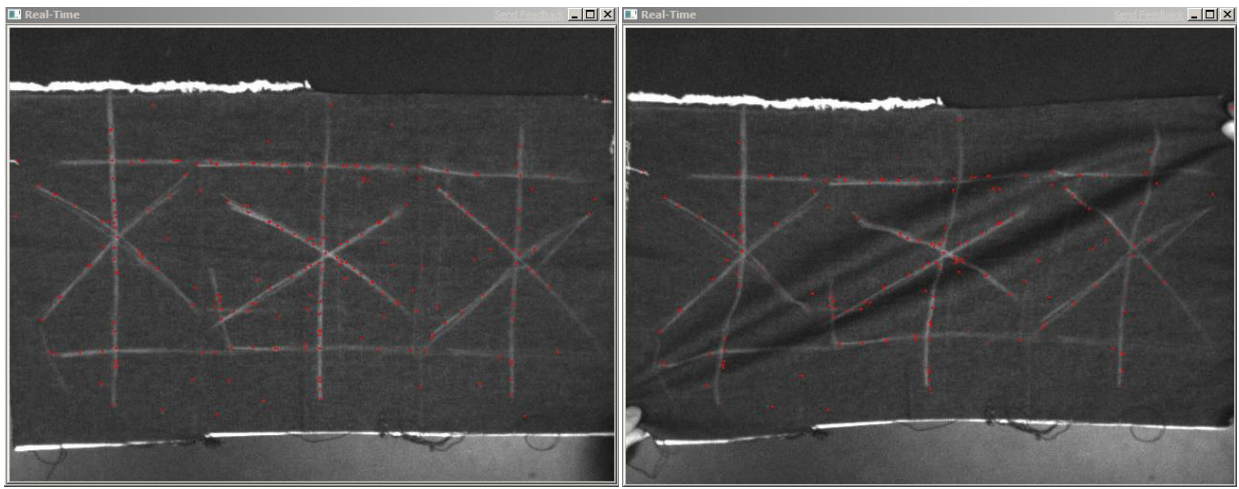

**Figure 19 Frame on the left shows initial features that were extracted, frame on the right shows features after cloth has deformed.**

Upon further testing, however, the features seemed to drift over time and large-scale translation. In order to test if this was a problem with the algorithm or with the LabVIEW setup, initial feature locations were extracted and saved for later use. Offline, the same frames were run with the given initial features. One can see in figure Figure 20 that the initial frames and features for the two methods are indeed the same.

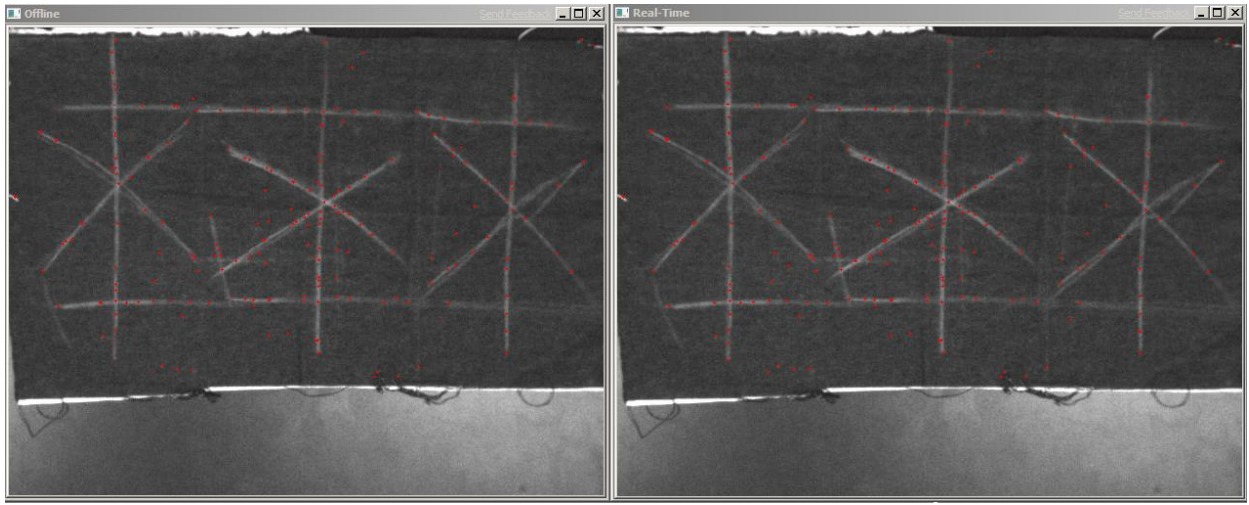

**Figure 20 Starting frames for offline and real-time feature tracking test, (on the left is the offline image, on the right is the real-time image).**

However, it is clear from Figure 21 that the features in the real-time frame on the right have begun to drift. By running the exact same frames with the same initial features offline, it is clear that the error in the real-time feature tracking is not with the OpenCV code.

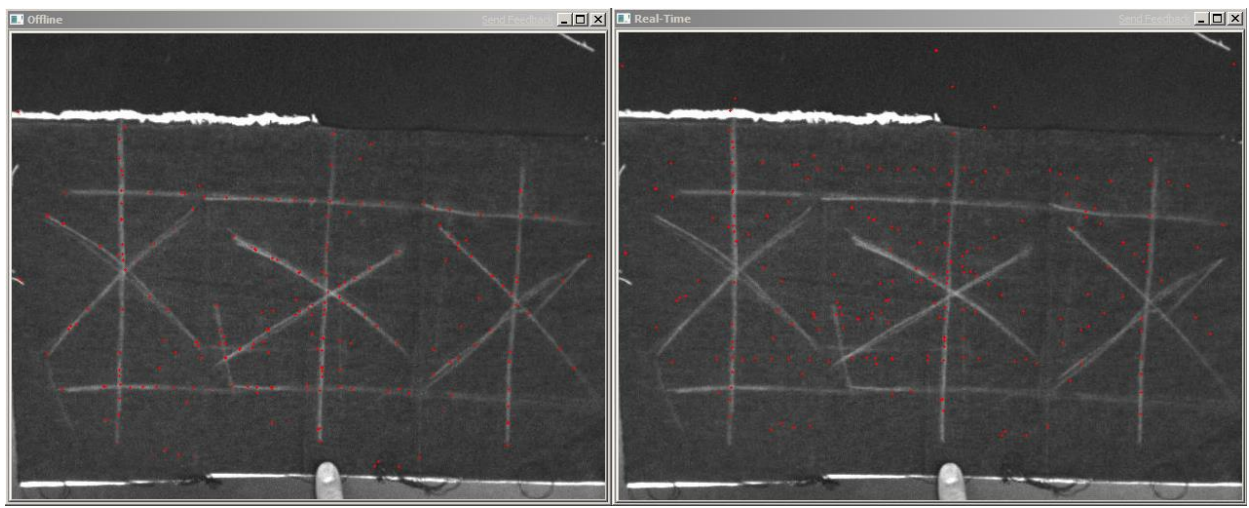

**Figure 21 Frames of offline and real-time tracking test after some displacement of the cloth (on the left is the offline image, on the right is the real-time image).**

The probable cause of this error is either network latency or errors in the real-time timing loop implementation in LabVIEW. The two processors on the host machine are only ever running at about 20% capacity during the experiment, which means that it is not likely an issue of processor power. Interestingly enough, even when the frames of the video sequence are processed offline on a quad-core 2.33 GHz desktop, the average processing time per frame was the same as the processing rate of the frame in the real-time implementation (around 15 fps). This likely means that the limiting factor on the speed of the complete cloth tracker is the speed of the 2D point tracker and not the network latency. The most likely cause for this error is unintentional asynchronous processing of frames due to errors in timing loops in the LabVIEW code.

## Conclusions and Future Work from Part 2

Although the tracker has not been fully integrated, both of the limiting issues from Part 1 have been addressed. First, a more robust, quick and accurate cloth model has been developed and proven useful in initial testing. Secondly, a real-time 2D point tracker has been implemented with minor errors and proof that the tracker itself is capable at tracking robustly at a fairly high frame rate.

Future work includes tying together existing components and reducing error. Of particular importance is the debugging of the 2D point tracker on the real-time system shown in Figure 18. In addition, tuning of parameters of the Extended Kalman filter including covariance matrices for the process and measurement are still required. Finally the Extended Kalman filter will need to be modified for every feature that is dropped and added from frame to frame. One option is to drop the number of points tracked every time a feature is lost until a critical level is reached and then resample the image for more features. The other option would be to resample certain regions on the fly when a number of features have been lost or the quality of those features has degraded. Finally, the forces applied to the cloth by conveyor motion will need to be estimated and serial communication with the stepper motors will need to be incorporated into the LabVIEW real-time system that has already been developed.

The experiments and implementations as well as future work will allow the validation and development of a fully automatic non-rigid cloth tracker. This tracker which will be able to run in real-time will allow the control of non-rigid cloth on a system with steerable conveyors.

# Related Work

**Cloth model**

Past work in simulating or modeling cloth behavior can be largely divided into geometrical or physical techniques or a hybrid of both. Ng et al. [48] summarized the initial work in this field. They explained that geometrical techniques do not consider properties of cloth but focus on appearance, especially folds and creases represented by geometrical equations. Physical techniques generally use triangular or rectangular grids with points of finite mass at the intersections.

The majority of physically based modeling techniques since the work described by Ng et al. [48] has been based around the methods developed by Baraff et al. in [3]. This work addressed the limitation of using small time steps to avoid numerical instability in numerical integration. Using the common mass-spring model, with a triangular mesh, they used a new technique for enforcing constraints on individual cloth particles with an implicit integration method. This meant that their method was always stable even for large time steps. Although much work since then has modified the method of constraints or the method of solving the linear system, their work was the seminal work in terms of cloth simulation using implicit integration. Desbrun et al. [19], [41] addressed the speed of Baraff's algorithm [3] by modifying the original implicit integration method in different ways. In [41] the authors used a hybrid explicit-implicit method in order to implement a stable real-time animation of textiles still based on a mass-spring model for a VR environment. Although it ran at a high frame rate its accuracy in comparison to real world cloth dynamics was never addressed. Work by Bridson et al. [10] is fairly representative of work that followed Barraf et al. [3] which focused on more realistic wrinkles or folds instead of speed of computation. In addition to using an explicit-implicit time integration scheme and mass-spring model, they introduced a physically correct bending model to better model wrinkles.

Work by Provot [50] uses the general mass-spring model and explicit integration techniques as well as limiting the elastic deformation arbitrarily when there is high stress in small areas. This allows for a more realistic modeling of real cloth without having to increase spring stiffness between the masses. Rudomin et al. [55] represent pieces of clothing using a mass-spring particle system implemented in real time. They point out that Euler or Runge-Kutta methods are fast and easy to implement, but require small steps. Whereas implicit integration can take much larger time steps but requires more computations to resolve the system at every step. According to their work, real time applications, where the time step is small and the regularity of the system cannot be determined, are best addressed using an explicit first or second order integration method. Bridson et al. [11] use the mass-spring model by Provot as the basis for an algorithm that deals robustly with collisions, contact and friction in cloth simulation by focusing on the internal dynamics of the cloth.

Another strain of work in the cloth modeling field that is much less common but is finding more popularity due to an increase in computational power and efficient formulations is that of a finite element formulation for deformable bodies. Eischen et al. [20] present a fairly representative survey of finite element methods for cloth modeling and develop their own algorithm based on nonlinear shell theory. Their motivation was to successfully simulate fabric drape and manipulation for use in textile and apparel manufacture. Most similar finite element methods were fairly slow in terms of run-time until the work of Etzmuss et al. [21]. They sped up the process by reducing the nonlinear elasticity problem to a planar linear problem for each implicit time integration step. Their algorithm was able to compute frames for a 0.02 second time step at an average rate of three seconds per frame for a simple shirt simulation and 16-21 seconds for a man walking in a shirt and trousers. Although still not real-time, this was a vast improvement on previous finite element methods with apparently little loss in accuracy although they include no proof of error measurement or validation. One other important and interesting point that they made is that from a finite element stand point, spring-mass systems should only use quadrilateral meshes as triangle meshes tend to show a larger shear resistance than real textiles. Their main criticism of work like Provot's was not its lack of accuracy compared to their algorithm but difficulty in applying it to garment construction for linking multiple pieces of simulated cloth. Finally, Garcia et al. [25] were able to sacrifice a very small amount of accuracy in order to make a finite element formulation for deformable bodies run at near 30 frames per second. They did this by estimating an initial solution and then only iterating on equations showing large error.

A succinct and current summary of the advantages and disadvantages among the different currently used methods of cloth simulation is found in work by Nealen et al. [47]. They discuss specifically the current deformable models that are all physically based, giving numerous examples at the end of their work.

**2D point tracking**

Among the numerous forms of tracking, (i.e. kernel tracking, contour tracking, active shape models, snakes), the most useful type for this specific application is point tracking. As the cloth's state will be represented using nodes, 2D feature tracks provide an obvious way to relate the motion of the feature points to the motion and velocity of the nodes of the cloth.

*Feature points*
The first step in feature tracking is being able to identify interest points in initial frames of the image or video sequence. An effective feature tracker therefore obviously depends on identifying "good features." Generally, feature point detectors that have proven effective in other applications are those that are insensitive to illumination variance and affine transformations. Mikolajczyk et al. [42][45] compared numerous scale and affine invariant point detectors and concluded that the SIFT detector by Lowe [34] performed the best in matching tests where images underwent affine and scale transformations. When the performance measure included changes in illumination, defocus and image compression as well they concluded that the MSER feature detector by Matas et al. [38] performed the best, with the Hessian affine detector being the next best [43]. They also concluded that for point tracking applications with occlusion or clutter, the Harris and Hessian affine feature detectors [43][45] were the most useful as they extracted more overall features than the other detectors. In work by Tissainayagam et al. [64] a number of different feature detectors ([32][67][62][43]) were implemented in order to track extracted points and compare the performance of the feature finders. It was shown that the two most effective types of feature points for tracking are Harris affine feature points [43] and KLT points (which were developed specifically for tracking purposes, see [67] and [60]). Some of the most recent work in feature extraction is focused on the speed and efficiency of extraction. Rosten and Drummond [54] (FAST) use machine learning to derive a feature detector that can run in real-time and is very competitive with other robust feature extractors. We will therefore be using KLT points or FAST feature points in our application since the SIFT feature extractor is generally much slower in running time than the other two methods.

The second aspect of point tracking is efficiently and effectively matching these feature points through progressive frames in order to produce 2D image tracks that can be used to update the estimate of the states from the cloth model. Many methods for this data correspondence problem have been proposed. Included here is a brief summary of the general areas and trends in feature tracking in relation to solving the data association problem.

*Kalman filtering*
Broida et al. [12] used a Kalman filter to track an object in noisy images. While Rosales et al. [53] used an extended Kalman filter in order to estimate a bounding box for position and velocity of multiple points which were then used in a larger scheme to estimate relative 3D motion trajectories. The major drawback with using a simple Kalman filter is that even though it provides optimal state estimates, it is only optimal for unimodal Gaussian distributions over the state to be estimated. In addition, the Kalman filter by itself does not solve the data correspondence problem. The authors Forsyth and Ponce [23] present a simple example of using the Kalman filter with a global nearest neighbor approach to solve the data correspondence problem.

*JPDA*
The joint probability distribution association (JPDA) addresses the data correspondence problem by computing a Bayesian estimate of the correspondence between features detected by the "sensor" and the different objects to be tracked. Fortmann et al. [24] first introduced this method with an application to a passive sonar tracking problem with multiple sensors and targets. The algorithm was also reviewed and evaluated by Cox [17] and a simple version of it is presented by Forsyth and Ponce [23]. The original JPDA still uses a Kalman filter and therefore is only optimal for Gaussian distributions over the state to be estimated. In addition, the JPDA algorithms fails to handle objects entering or exiting the scene. This is particularly a problem because the JPDA only associates the data over two frames which can lead to major errors if the number of objects changes. Schulz et al. [58] addressed the limiting factor of the JPDA only describing a Gaussian probability distribution function by presenting a sample-based JPDA filter to track multiple objects. This method then combines the advantages of sample-based density approximations with the generally efficient JPDA. This does not however resolve the problem of objects entering or leaving the scene or at least being occluded in the two frame interval. Because the JPDA is not formulated to handle objects entering or leaving the scene and because for point tracking, sample-based density approximations seem irrelevant, this method is a less likely candidate for a cloth control problem.

*MHT*
As an alternative to the JPDA method of data correspondence, Reid [52] proposed the multiple hypothesis tracking algorithm (MHT). The underlying principle of the MHT algorithm is maintaining multiple hypotheses about a single object in order to obtain optimal data correspondence. He also incorporated the ability to initiate tracks, account for false or missing objects and process sets of dependent reports in one algorithm. One major drawback is the obvious combinatorial explosion that takes place as all hypotheses are explored. Cox et al. [16][18] updated the original MHT algorithm to run more efficiently and applied it to feature point tracking. They tracked the features using a simple linear Kalman filter and kept only the k-best hypothesis. They used Murti's algorithm [46] to solve the problem and additionally addressed track initiation, termination and low-level support for temporary track occlusion. This work has been used more recently by Tissainayagam et al. [63][65]. The authors use the MHT method twice in a framework where they both segment contours and track selected key points using the MHT algorithm for both parts.

However they make no claim about being able to run in real-time which is clearly important in being able to use the tracking in any kind of a control application. In addition, the authors explain that the tracking process can breakdown on the contour segmentation side for occlusion or possibly deforming objects which is clearly a problem as our objective is to track and control non-rigid cloth. Veenman et al. [70] have shown that tuning parameters for the MHT algorithm tends to be tricky as it is extremely sensitive to its parameter settings. Finally, the complexity of the algorithm still grows exponentially with the number of points making the MHT algorithm a less effective feature tracker in non-rigid tracking applications which are facilitated by a larger number of 2D point tracks.

*Deterministic Algorithms*
Deterministic algorithms are another class of feature trackers that have seen quite a bit of success using qualitative motion heuristics instead of probability density functions and distributions. Salari et al. [56] formulated the data correspondence problem as a minimization problem for extracting globally smooth trajectories that are not necessarily complete but satisfy certain local smoothness constraints. Their work involved an iterated optimization procedure. They were followed by the work of Rangarajan et al. [51] who defined a proximal uniformity constraint which said that most objects in the real world follow smooth paths and cover small distances in a small amount of time. They also minimized a cost function that described such a constraint and used gradient based optical flow to establish correspondence in the initial frames. Their algorithm was a non-iterative greedy algorithm. Veenman et al. [70] built on this previous work and developed what they call a qualitative motion modeling framework. Their algorithm was less sensitive to parameter changes than the previous deterministic algorithms and outperformed them as well as outperforming the most recent MHT algorithms at the time despite running in real-time. They also included error and occlusion handling and did automatic initialization of tracks. The limitations however were that this method addressed no track initiation or termination during the image sequence and optimized only over two frames. Shafique et al. [59] developed another non-iterative greedy algorithm that presented a framework for finding the correspondence over multiple frames using a single pass greedy algorithm. Their method handles object entry or exiting of the frame as well as occlusion. Their results show that the algorithm can run at 17 Hz for the tracking of 50 separate points on a 2.4 GHz Intel Pentium 4 CPU although they do not mention for what resolution this measurement is given. Their method automatically initializes the tracks using the Shi-Tomasi-Kanade tracker [60]. Finally showing that the algorithm of Shafique et al. was useful in tracking non-rigid objects, Mathes and Piater [39] developed 2D point distribution models using the automatically tracked points from [59] to track non-rigid motion of football (soccer) players. The real-time implementation of these deterministic algorithms as well as their current application in other non-rigid tracking problems makes them a candidate for our application.

*Optical flow estimation*
One of the most popular methods for point tracking is based on the two-frame differential method for optical flow estimation. Lucas and Kanade [35] first presented an iterative approach that measured the match between fixed-size feature windows in past and current frames as the sum of squared intensity differences of the windows. The track or correspondence was then defined as the one that minimized the sum between two frames. Tomasi and Kanade [67] re-derived the same method in a more intuitive fashion and then addressed how to define feature windows that were best suited for tracking. Shi and Tomasi[60] continued this work by proposing a feature selection criterion based on how the tracker works so as to pick "good features" for this tracking method. They also proposed a feature monitoring method (i.e. better addressing the correspondence problem) that can handle occlusions, disocclusions and features that do not correspond to points in the real world (noise). Finally they extended previous Newton-Rapshon style search methods to work under affine image transformations. Their work was extended by Tommasini et al. [68] who introduced an automatic scheme for rejecting spurious features by using a simple outlier rejection rule. Zinsser et al. [73] proposed two ameliorations to the Shi-Tomasi-Kanade tracker which dealt with affine motion estimation and feature drift. In addition their algorithm ran in real-time and was specifically formulated to run well over long image sequences. Sinha et al. [61] implemented the KLT tracker [67] on the graphics processing unit (GPU) instead of the CPU. Two examples of work that have successfully used the implementation by Sinha et al. is the work by Akbarzadeh et al. in 3D urban reconstruction [2] and Andreasson et al. in a real-time SLAM application [1]. They were able to track about a thousand features in real-time at 30 Hz on 1024x768 resolution video. An alternative to the implementation of the Lucas Kanade tracker by Sinha is the pyramidal method developed by Bouguet [6] and implemented in the computer vision library OpenCV [9]31. Because of the wide use of this approach and its implementation in real-time, this method is one of the best candidates for being able to track the non-rigid cloth.

**Cloth tracking**

Non-rigid tracking is not a new topic in research and is fairly well developed for some applications. However, algorithms such as the one found in the work by Comaniciu et al. [14] is not acceptable for a cloth tracking application. This is because only a bounding box or contour of the non-rigid object is tracked which is useful for a cloth tracking application since we want to track and control within the perimeter of the cloth as well.

In terms of actual tracking or parameter estimation of cloth from real video, most current methods use implicit integration for their cloth model. Bhat et al. [4] used the general method proposed by Baraff et al. [3] to present an algorithm for estimating the parameters for cloth simulation from video data of real fabric. Simulated annealing was used to minimize the frame by frame error between a given simulation and the real-world footage. Pritchard et al. [49] instead used the fairly well developed methods of rigid body motion capture systems to estimate the geometry of a non-rigid textile. They pointed out that methods proposed by Bridson and Baraff are still too demanding for real-time applications. Instead, they proposed using feature matching techniques and a multi-baseline stereo camera setup with 10fps to recover geometry and parameterize a moving sheet of cloth. Scholz et al. [57] also used a multi-camera setup but used optical flow between frames to model the deformable surface. They included a silhouette matching procedure which is required to correct the tracking errors that occur in long video sequences.

Similar work was done by Hasler et al. [24][25] in terms of cloth parameter estimation for a tracking application using analysis-by-synthesis. Their method consisted of optimizing a set of parameters of a mass-spring model that were used to simulate the textile. The fabric properties and the positions of a limited number of constrained points of the simulated cloth were found during the optimization. However, in order to improve tracking accuracy, nonphysical forces were introduced to bias the simulation towards following the observed real world behavior. Due to the optimization scheme used and the fact that they estimated all parameters for every frame, their algorithm took 10 to 30 hours on 7 AMD Opteron processors at 2.2GHz to converge for their experiments. Where the work by Hasler et al. was based on surface features of the cloth, White et al. [72] use painted markers of constant color. They used multiple cameras which allowed them to compute the markers' coordinates in world space using correspondence. They also addressed a data driven hole-filling technique for occluded regions by using previous frames where the occluded regions were visible.

One area of cloth tracking that has been steadily growing is the use of garment capture for use with augmented reality, post production of films or even avatars. Bradley and Roth [7] were able to track non-rigid cloth and insert virtual images on the non-rigid cloth in real-time. However, they required special patterns in specific locations to be on the cloth. In similar work, Bradley et al. [8] were able to successfully capture complicated garment motion with a sixteen camera array and then insert images or change the clothing completely in post-production. The disadvantages were the complicated setup and the fact that processing times for each frame were approximately one hour. Another approach to the same problem was presented by Hernandez et al. [28] who used multispectral photometric stereo to recreate 3D data for a moving non-rigid object. Although they obtain impressive results, their processing time of 40 seconds per frame for a 1280x720 resolution image with a 2.8 Ghz processor is too slow for a control application. In addition the method is perhaps too restrictive for textile manufacturing in requiring specific lighting conditions that may not be possible in a factory setting.

**Visual tracking and control**

Recent work on combining the aspects of visual tracking and control focuses on robustness and speed. Work by Comport et al. [15] developed a 3D model-based tracking method with a focus on trying to more robustly handle natural scenes without fiducial markers. In [37], Marchand et al. gave a review of effective 2D feature-based or motion-based tracking methods that have been used over the past 10 years in addition to reviewing their formulation of the 3D model-based tracking method. Malis et al. [36] developed a combined visual tracking and control system similar to our objective. However, they proposed a template matching algorithm based on second-order minimization in an effort to avoid design of feature dependent visual tracking algorithms. Although clearly important for more broad applications, template based tracking is inadequate to be able to measure the states of a non-rigid textile to be tracked. Kumar et al. [33] developed visual servoing strategies that involved robustness to non-rigid deformation. However, the non-rigid deformation is assumed to be part of the environment of the robot and not the object to be controlled per se. Finally, Tran et al. [69] developed a tracking algorithm for visual servoing which most resembles our proposed plan. However, even using a fast corner detector, a corner descriptor based on Principal Component Analysis and an efficient matching scheme using approximate nearest neighbor technique they were only able to achieve tracking at 10-14 fps. This is sluggish in terms of desired control rates.

**Stepper motors and control**

Although both servo and stepper motors are a possibility for controlling the cloth on an array, stepper motors offer a generally more accurate solution. The behavior and capabilities of stepper motors have been extensively studied. Numerous control schemes including linear and non-linear methods have been proposed , [74][40][5]. Frequency response, stability and nonlinear behavior of stepper motors were studied by Cao and Schwartz [13]. Most recently, Ferrah et al. [22] presented a clear and simple stepper motor model and developed an extended Kalman Filter to improve the control over the already accurate stepper motor.

Although much work has been done in terms of visual servoing and tracking of rigid objects, tracking and control of a non-rigid object like cloth has not been addressed. It can be seen from the review above that combining techniques from the well-developed field of cloth simulation as well as the field of feature tracking will allow us to further develop work in real-time textile tracking as well as textile control.